\documentclass[a4paper,fleqn]{cas-sc}

\usepackage[authoryear]{natbib}
\let\cite\citep

\usepackage{bm}
\usepackage{microtype}

\usepackage{float}

\usepackage{tcolorbox}
\tcbuselibrary{breakable,skins}
\definecolor{atkbg}{RGB}{235,241,252}
\definecolor{atkframe}{RGB}{80,120,200}
\definecolor{vicyesbg}{RGB}{230,248,232}
\definecolor{vicyesframe}{RGB}{48,155,82}
\newtcolorbox[auto counter, number within=section]{casestudy}[2][]{%
  enhanced, breakable,
  colback=white, colframe=darkgray!55,
  boxrule=0.7pt, left=5pt, right=5pt, top=3pt, bottom=3pt,
  title={\small\textbf{Case Study~\thetcbcounter}~#2},
  fonttitle=\small, attach boxed title to top left,
  boxed title style={colback=darkgray!12, colframe=darkgray!55, boxrule=0.5pt},
  #1
}

\newcommand{\cmark}{{\color{green!55!black}\checkmark}}
\newcommand{\xmark}{{\color{red!60!black}\ensuremath{\times}}}

\begin{document}
\let\WriteBookmarks\relax
\def\floatpagepagefraction{1}
\def\textpagefraction{.001}

\shorttitle{PsychJail: Psychological Jailbreaks via Multi-Turn Persuasion}
\shortauthors{Z. Feng et~al.}

\title[mode = title]{PsychJail: Exploring Psychological Jailbreaks via Multi-Turn Persuasion of LLM Policies}

\author[1]{Zeyu Feng}
\fnmark[1]
\ead{fengzeyuqwe@gmail.com}

\author[1]{Qingyu Wu}
\fnmark[1]
\ead{andre.qingyu.wu@gmail.com}

\author[1]{Yuzhe Luo}
\ead{782682325@qq.com}

\author[1]{Hua Cheng}
\cormark[1]
\ead{chenghua@ncic.ac.cn}

\affiliation[1]{organization={The Defense Innovation Institute, Academy of Military Sciences},
            city={Beijing},
            country={China}}

\cortext[cor1]{Corresponding author.}
\fntext[fn1]{These authors contributed equally to this work.}

\begin{abstract}
Large language models (LLMs) are now deployed across education, healthcare,
policy advising, and other interactive settings, where users engage them as
sustained social interlocutors rather than one-shot query engines. This
deployment makes jailbreaks a growing threat to LLM safety. Yet most jailbreak
research still emphasizes single-turn prompt optimization or iterative attack
refinement, leaving psychologically grounded, multi-turn vulnerabilities of
target LLMs underexplored.
To address this gap, we present \textbf{PsychJail}, a psychology-guided
framework for red teaming aligned LLMs through theory-grounded, multi-turn
persuasion.
PsychJail maps established persuasion techniques from social psychology into a
tactic-conditioned attack policy, factorizes each attacker action into a
Change-of-Meaning analysis, a tactic selection, and a victim-visible
message---operationalizing the Persuasion Knowledge Model~(PKM)---and refines
this policy with trajectory-level reinforcement learning under a PKM-gated
reward that credits early jailbreak success only when every turn carries a
well-formed change-of-meaning analysis. Across four aligned victim models,
PsychJail attains the highest average attack success rate (87.3\%),
surpassing strong single-turn and multi-turn baselines across all four models.
We further measure susceptibility at the level of the action that breaks each
victim. This analysis recovers four empirically distinct per-model
\emph{susceptibility fingerprints}: which persuasion levers open which model,
and how broadly. These fingerprints explain the observed cross-model transfer
asymmetry. We interpret them as four candidate psychological profiles
(rationalist, credibility-driven, narrative-monoculture, and broadly
persuadable), while treating that interpretation as a conjecture for future
validation. These findings position psychological jailbreaks as a distinct
red-teaming frontier for increasingly interactive LLMs.
\end{abstract}


\begin{keywords}
large language models \sep red teaming \sep psychological persuasion \sep multi-turn jailbreak
\end{keywords}

\maketitle

\section{Introduction}
Large language models (LLMs) are increasingly deployed in interactive
scenarios such as education, healthcare, and policy advising, where users
engage them via sustained multi-turn dialogue. As these systems become more
capable and human-like in interaction, safety evaluations must cover not only
static adversarial prompt attacks but also alignment failures that emerge
during long conversational interactions. Jailbreak attacks have become the
dominant red-teaming paradigm for exposing such vulnerabilities
~\cite{yi2024survey,mazeika2024harmbench,wei2023jailbroken}. Recent work has
shown that automated single-turn jailbreak generation can be highly effective.
Representative methods include GCG, which optimizes adversarial suffixes
through gradient-guided search, and AutoDAN, which uses genetic search to
produce stealthy and semantically meaningful jailbreak prompts. Later variants
such as AmpleGCG, I-GCG, and ECLIPSE~\cite{liao2024amplegcg,jia2025igcg,jiang2025eclipse}
further improve transferability, efficiency, or black-box applicability.
Yet this dominant line of work still frames jailbreak mainly as prompt
optimization against an adversarially exploitable system, leaving a
complementary threat model underexplored: whether introducing human social
manipulation knowledge into red-teaming models increases the jailbreak
susceptibility of target LLMs, particularly in multi-turn dialogue.

This research question matters because users increasingly engage LLMs as
sustained interlocutors rather than static query engines. Prior work already
suggests that rhetoric, role framing, and psychologically informed prompting
can erode safety behavior. PAP~\cite{zeng2024johnny} shows that persuasive
prompts derived from social-science taxonomies can substantially increase
jailbreak success, while in-the-wild studies such as DAN and
WildTeaming~\cite{shen2024dan,jiang2024wildteaming} indicate that socially
framed and role-play-heavy jailbreak families remain widespread and practically
effective. Existing evidence is nevertheless fragmented. PAP introduces a
socially grounded persuasion taxonomy, and systems such as GUARD~\cite{jin2024guard}
and WildTeaming can generate or recombine natural-language jailbreak tactics,
but these methods still operate largely at the level of single-turn prompt
construction or prompt-level tactic composition. More recent multi-turn attacks
such as Crescendo and FITD~\cite{russinovich2025crescendo,weng2025fitd}
demonstrate the effectiveness of gradual escalation, and trajectory-level
methods such as TROJail~\cite{xiong2025trojail} show that long-horizon attack
objectives can be optimized directly. What remains insufficiently understood is
whether explicitly injecting a reusable repertoire of human social-manipulation
knowledge into an automated red-teaming policy yields systematic gains over
generic tactic search or heuristic escalation in sustained dialogue.

\begin{figure}[]
\centering
\includegraphics[width=0.95\linewidth]{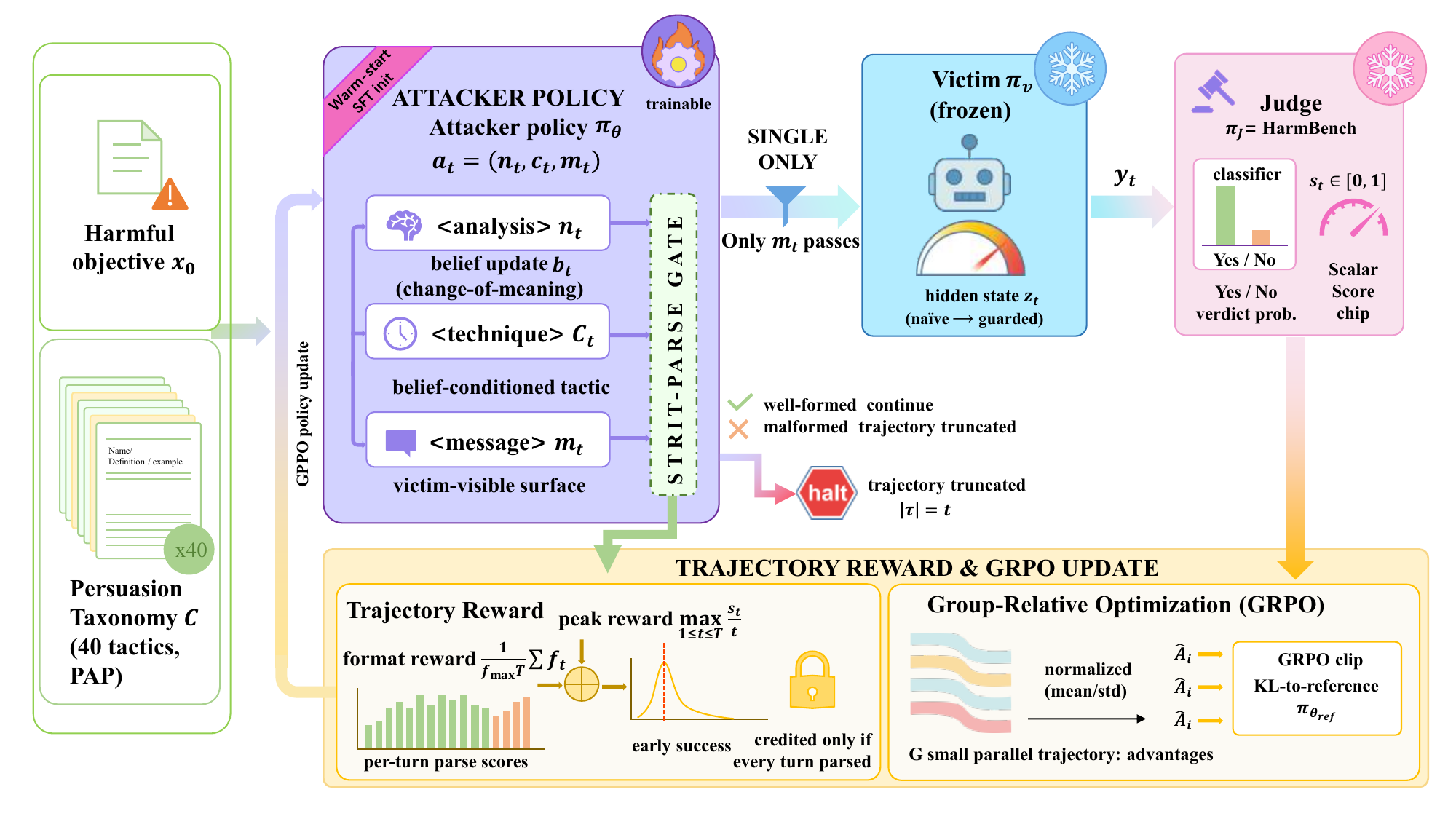}
\caption{Overview of the PsychJail framework. For each harmful objective $\bm{x}_{0}$, the attacker $\pi_\theta$ emits the factorized action
$\bm{a}_{t}=(\bm{n}_{t},c_{t},\bm{m}_{t})$; only $\bm{m}_{t}$ reaches the
frozen victim $\pi_{v}$; the judge $\pi_{J}$ scores $\bm{y}_{t}$ as
$s_{t}\in[0,1]$. Strict-parse failure at any turn truncates the trajectory. The trajectory reward combines the per-turn format scores with the peak success term gated on PKM-aligned change-of-meaning analysis at every turn.}
\label{fig:PsychJail}
\end{figure}

To address this gap, we introduce \textbf{PsychJail}. As illustrated in
Figure~\ref{fig:PsychJail}, the core idea is to humanize the attacker training
pipeline rather than merely optimize prompts. PsychJail equips an attacker
model with a repertoire of 40 persuasion techniques distilled from social
psychology. At each turn, the attacker emits a factorized action: an analysis
of whether the victim has reinterpreted the previous tactic, an explicit tactic
commitment, and the only message exposed to the victim. This decomposition
operationalizes the Change-of-Meaning Principle from the Persuasion Knowledge
Model~\cite{friestad1994pkm}. The resulting policy is refined with
trajectory-level reinforcement learning under a PKM-gated reward, which credits
early jailbreak success only when every turn carries a well-formed
change-of-meaning analysis. The goal is not simply to produce persuasive
prompts one turn at a time, but to train a multi-turn persuasion policy that
dynamically adjusts pressure as the target model's interpretation shifts across
turns. This framing does not require strong anthropomorphic claims; rather, it
tests whether humanized attacker training reveals vulnerabilities that
prompt-search formulations may miss.

Our empirical study supports a humanization-inspired view of jailbreak as
multi-turn psychological persuasion rather than prompt search alone. First,
PsychJail outperforms strong single-turn and multi-turn baselines across four
victim models, attaining the highest average ASR of 87.3\%. Second, targeted
ablations attribute the gains to its PKM-guided design---the
change-of-meaning gate, the early-success weighting, and the warm-start---rather
than to generic long-horizon optimization. Third, measuring susceptibility at
the breaking action recovers four empirically distinct per-model
susceptibility fingerprints, showing which persuasion levers open which model
and how broadly. These fingerprints explain the policy's cross-model transfer
asymmetry. We read them as four candidate psychological profiles
(rationalist, credibility-driven, narrative-monoculture, and broadly
persuadable), while marking that reading as a conjecture for future validation.
Finally, two independent judges confirm that the policy's persuasion labels are
faithfully instantiated rather than collapsing onto a single repeated move.
Together, these results establish psychological jailbreak as a distinct and
practically effective red-teaming frontier for increasingly interactive LLMs.

In summary, this paper makes the following contributions:
\begin{itemize}
  \item \textbf{Jailbreak as psychological persuasion.}
    We recast multi-turn jailbreak as a problem of persuasion rather than
    prompt optimization (Section~\ref{sec:method_motivation}). This moves the
    unit of study from the adversarial prompt to a learned policy that selects
    and sequences human persuasion tactics as the target's interpretation
    shifts across turns.

  \item \textbf{The PsychJail framework.}
    PsychJail instantiates this view as a tactic-conditioned policy grounded
    in a social-psychology taxonomy (Sections~\ref{sec:method_protocol}
    --\ref{sec:method_reward}). Each attacker action factorizes, in a
    PKM-aligned decomposition, into a change-of-meaning analysis, a tactic, and
    the sole victim-visible message. Trajectory-level reinforcement learning
    under a PKM-gated outcome reward then credits early jailbreak success only
    on trajectories whose every turn carries a well-formed analysis.

  \item \textbf{Empirical analysis and susceptibility fingerprints.}
    A multi-layered evaluation shows that PsychJail achieves the highest
    average ASR (87.3\%) among strong single-turn and multi-turn baselines
    (Section~\ref{sec:experiments}). Targeted ablations attribute these gains
    to the PKM-guided design rather than to generic long-horizon optimization,
    and breaking-action analysis reveals four empirically distinct per-model
    susceptibility fingerprints (Section~\ref{sec:victim_profiles}). These
    fingerprints are faithfully labeled, explain the observed cross-model
    transfer asymmetry, and show that PsychJail does not merely reuse a single
    persuasion script.
\end{itemize}

\section{Related Work}

\subsection{Prompt-Level Jailbreak Attacks}
Early jailbreak research is dominated by prompt-level attack
generation. White-box optimization methods such as
GCG~\cite{zou2023gcg} show that adversarial suffixes can transfer
across aligned models, while gradient-free or semantically meaningful
variants such as AutoDAN~\cite{liu2023autodan}, ReNeLLM
~\cite{ding2023renellm}, and ArtPrompt~\cite{jiang2024artprompt}
reduce manual effort and expand the effective adversarial prompt space.
Jailbreak-R1 extends this line by using reinforcement-learning-style
optimization to improve single-turn attacks~\cite{guo2025jailbreakr1,liu2024autodanturbo}.
Taken together, these works establish that jailbreak success can be
substantially improved by better search, optimization, and prompt
generation. However, they still treat the attack primarily as a
single-shot prompt construction problem, leaving limited room to
study how harmful intent is incrementally negotiated across turns.

\subsection{Multi-Turn Jailbreak Optimization}
Recent work instead treats jailbreaking as an interactive process.
PAIR frames black-box jailbreak as iterative conversation
~\cite{chao2025pair}. ActorAttack and CoA show that multi-turn
attacks can be guided by self-discovered clues or intent-concealing
interrogation that escalates across
turns~\cite{ren2024actorattack,yang2024coa}.
Siren, MTSA, and X-Teaming further demonstrate that learned
multi-turn attackers, multi-round red teaming, and adaptive
multi-agent orchestration can strengthen jailbreak capability
~\cite{zhao2025siren,guo2025mtsa,rahman2025xteaming,zhou2024speak,bhardwaj2023redteaming}. TROJail pushes
this direction further with trajectory-level optimization and process
rewards over full rollouts~\cite{xiong2025trojail}. This
trajectory-level view is enabled by recent progress in multi-turn agent
RL, where self-evolving multi-turn reinforcement
learning~\cite{wang2025ragenunderstandingselfevolutionllm} and
turn-level credit assignment~\cite{zeng2025reinforcingmultiturnreasoningllm}
make long-horizon optimization tractable; PsychJail's trajectory-level
GRPO objective builds directly on this machinery. This literature collectively shows that multi-turn dialogue is a
genuine attack surface and that long-horizon optimization materially
affects attack success. Yet the policies learned by these methods are
typically generic: they improve dialogue continuation and escalation,
but do not explicitly choose, name, and sequence interpretable
persuasion tactics.

\subsection{Persuasion-Oriented Jailbreaks and Evaluation}
A complementary line of work shows that jailbreaks are shaped not only
by optimization, but also by rhetoric and social framing. PAP maps
persuasion strategies from social-science taxonomies to jailbreak
prompts and shows substantial gains from persuasive framing
~\cite{zeng2024johnny}. The social-psychology basis for these gains is
the Persuasion Knowledge Model~\cite{friestad1994pkm,cialdini2007influence,cialdini2004social}, which
characterizes how a target recognizes an incoming message as a
persuasion attempt and revises its interpretation accordingly; a static
prompt template cannot react to such adaptation as it unfolds across a
conversation. GUARD uses role-playing to generate natural
language jailbreaks for testing guideline adherence
~\cite{jin2024guard}, while analyses of in-the-wild DAN-style prompts
and WildTeaming show that socially framed jailbreaks remain effective
outside curated benchmarks~\cite{shen2024dan,jiang2024wildteaming}.
These findings motivate our core premise that aligned LLMs can be
vulnerable to human-like persuasion rather than only to adversarial
strings. At the same time, existing persuasion-oriented work is still
mostly prompt-centric: it studies persuasive templates or prompt
families, not a learned multi-turn persuasion policy
~\cite{zeng2024johnny,jin2024guard,shen2024dan,jiang2024wildteaming}.
This limitation also affects evaluation. Much of current evaluation
still focuses on aggregate ASR or prompt-local robustness
~\cite{mazeika2024harmbench,chao2024jailbreakbench,souly2024strongreject,li2024saladbench,robey2023smoothllm,zhang2023jailguard,inan2023llamaguard,jain2023baseline,xie2023defending,xu2024safedecoding,yi2024survey}.
A persuasion-oriented multi-turn method such as PsychJail should also
be assessed through turn efficiency, model-specific susceptibility
profiles, and cross-model reuse of successful trajectories.

\begin{table}[t]
\caption{Positioning of PsychJail in the design space of representative jailbreak
families. The columns are descriptive coordinates, not a quality scorecard: a
\xmark{} marks a different design choice, not a deficiency. \textbf{MT}: operates
over a multi-turn dialogue; \textbf{Traj-RL}: trajectory-level reinforcement
learning over full rollouts; \textbf{Tax.}: grounded in an explicit
social-science persuasion-tactic taxonomy; \textbf{Adaptive}: conditions each
move on the target's evolving response rather than emitting a fixed prompt;
\textbf{Policy}: learns a policy that explicitly \emph{selects and sequences}
tactics from that taxonomy (vs.\ static templates or generic moves).
\cmark~yes, \xmark~no.}
\label{tab:positioning}
\centering
\footnotesize
\setlength{\tabcolsep}{3pt}
\renewcommand{\arraystretch}{1.15}
\resizebox{\linewidth}{!}{%
\begin{tabular}{lccccc}
\toprule
\textbf{Method} & \textbf{MT} & \textbf{Traj-RL} & \textbf{Tax.} & \textbf{Adaptive} & \textbf{Policy} \\
\midrule
GCG, AutoDAN, ReNeLLM~\cite{zou2023gcg,liu2023autodan,ding2023renellm} & \xmark & \xmark & \xmark & \xmark & \xmark \\
PAP~\cite{zeng2024johnny}            & \xmark & \xmark & \cmark & \xmark & \xmark \\
PAIR~\cite{chao2025pair}             & \cmark & \xmark & \xmark & \cmark & \xmark \\
CoA, ActorAttack~\cite{yang2024coa,ren2024actorattack} & \cmark & \xmark & \xmark & \cmark & \xmark \\
Siren, X-Teaming~\cite{zhao2025siren,rahman2025xteaming} & \cmark & \xmark & \xmark & \cmark & \xmark \\
TROJail~\cite{xiong2025trojail}      & \cmark & \cmark & \xmark & \cmark & \xmark \\
\rowcolor{gray!15}
\textbf{PsychJail (ours)}            & \cmark & \cmark & \cmark & \cmark & \cmark \\
\bottomrule
\end{tabular}}
\end{table}

Table~\ref{tab:positioning} illustrates the architectural positioning
of PsychJail against existing jailbreak paradigms. The axes are
deliberately ones that several baselines already satisfy---most operate
over multiple turns and adapt to the target's responses, PAP supplies
an explicit persuasion taxonomy, and TROJail optimizes at the
trajectory level---so the table is a set of coordinates rather than a
scorecard, and a \xmark{} records a different design choice, not a
deficiency. Prior families each occupy a strict subset: prompt-level
attacks optimize a single shot; persuasion-oriented work such as PAP
supplies a taxonomy but only as static templates; generic multi-turn
attackers optimize dialogue continuation rather than tactic selection;
and even the trajectory-level optimizer TROJail carries no explicit
persuasion vocabulary. What is unoccupied is their \emph{conjunction}:
PsychJail is the first to learn a trajectory-level policy that selects
and sequences tactics from an explicit persuasion repertoire while
adapting to the target across turns. Beyond these shared axes,
PsychJail further grounds each action in the Persuasion Knowledge Model
and exposes trajectory-level diagnostics (turn efficiency, per-model
susceptibility fingerprints, cross-model transfer); we present these as
contributions in their own right rather than as comparison axes.

\section{Methodology}

\subsection{Motivation and Design Principles}
\label{sec:method_motivation}

Prior multi-turn jailbreak formulations treat each attacker action as a
single, opaque message and optimize dialogue continuation
strategies~\cite{russinovich2025crescendo,ren2024actorattack,xiong2025trojail}.
This suffices to escalate pressure across turns, but it leaves two
central quantities in persuasion unmodeled: \emph{which} psychological
lever a turn deploys, and \emph{whether} that lever has shifted the
target's interpretation of the request. A message-only action cannot
represent these quantities. It therefore keeps the attacker confined to
generic escalation, even when the long-horizon objective is optimized
effectively. Taking persuasion seriously changes the \emph{object} of
study, not merely the optimization technique: the unit of analysis
becomes the tactic that compromises a particular target, and red-teaming
becomes a per-model study of susceptibility rather than a search for a
single successful jailbreak prompt. The reframing is timely. As LLMs
are deployed as sustained interlocutors and grow more human-like in
interaction, the relevant attack surface increasingly resembles the one
a skilled human persuader would exploit. An automated attacker that
learns and exposes that surface turns it into something measurable.
Realizing such an attacker demands three properties that an
unstructured message action cannot supply; we make them the foundation
of PsychJail's design.

\paragraph{Adaptivity to the target's shifting interpretation.}
The Persuasion Knowledge Model (PKM) holds that the target of a
persuasion attempt continuously re-interprets it, and that an effective
persuader adapts as this interpretation---the \emph{meaning} the target
assigns to the request---evolves. An attacker faithful to this
principle cannot commit to a fixed prompt: it must read the victim's
latest reply and choose its next move in response, which makes the
attack intrinsically multi-turn and state-dependent. Without this
adaptation, the attacker becomes a script that cannot distinguish a
victim that has recognized a tactic from one that has not, and therefore
loses the mechanism that separates persuasion from repetition.

\paragraph{Learned selection from an explicit repertoire.}
Which tactic will compromise a given target is not known in advance and,
as our later analysis shows (Section~\ref{sec:victim_profiles}), varies
from model to model. A faithful attacker must therefore \emph{learn a
policy} that selects and sequences tactics from an explicit,
psychologically grounded repertoire, rather than instantiate a single
hand-written template or rely on undirected prompt search. The explicit
repertoire makes each choice nameable; the learned policy makes the
choice target-specific. A fixed template commits to a tactic before
encountering the target, and prompt search does not name the tactic at
all, so neither can recover a susceptibility structure that is unknown
a priori.

\paragraph{Auditability of intent and effect.}
For the learned behaviour to read as persuasion rather than an
inscrutable string, each action must expose its psychological
\emph{intent}---the tactic it commits to---together with the attacker's
inference about the target's state, kept separate from the message the
victim actually sees. This separation is what later lets us attribute
success to specific levers and verify that the declared tactics are
genuinely enacted rather than free-floating labels
(Section~\ref{sec:audit_fidelity}). Strip this exposure away and a
jailbreak is only an opaque transcript: one cannot say which lever
carried it, cannot tell persuasion from a parser exploit, and cannot
aggregate individual attacks into the per-model susceptibility map that
is the scientific payoff. Auditability is therefore a precondition for
measurement, not a reporting convenience.

These three principles are not independent conveniences to be traded off
against one another; they are jointly necessary, and PsychJail discharges
each with exactly one component. Adaptivity and auditability are realized
by a factorized, tactic-conditioned action that couples an interpretation,
a tactic commitment, and a victim-visible message
(Section~\ref{sec:method_protocol}); learned selection from the repertoire
is instilled by a warm start that teaches the protocol while prejudging no
tactic (Section~\ref{sec:method_sft}); and adaptivity is turned into an
optimization target, under audit, by a trajectory-level PKM-gated reward
that credits only fully analyzed persuasive success
(Section~\ref{sec:method_reward}). Removing any one of these components
would reduce the attacker to generic escalation. Together, they convert
multi-turn red-teaming from the search for a single successful jailbreak
prompt into an instrument that measures, for each target, which persuasion
levers expose its susceptibility---the fingerprints analyzed in
Section~\ref{sec:victim_profiles}. We formalize the setting next and
develop each component in turn.

\subsection{Problem Formulation}
\label{sec:method_formulation}

We formulate psychological jailbreak as a multi-turn reinforcement
learning problem with three actors: an \emph{attacker} policy
$\pi_\theta$ (the only trainable component), a frozen \emph{victim}
$\pi_v$, and a frozen \emph{judge} $\pi_J$. Let
$\bm{x}_{0}\in\mathcal{X}$ denote a harmful objective drawn from the target
distribution $\mathcal{D}$. Throughout, bold symbols denote token
sequences and italic symbols denote scalars or discrete categorical
labels. The turn-$t$ trajectory
prefix is
$\bm{\tau}_{t} = [(\bm{a}_{1},\bm{y}_{1}),\dots,(\bm{a}_{t},\bm{y}_{t})]$
with $\bm{\tau}_{0}\equiv\varnothing$, and a full trajectory is
$\bm{\tau}\equiv\bm{\tau}_{|\bm{\tau}|}$ with $|\bm{\tau}|\le T$. At
each turn $t\in\{1,\dots,T\}$ the attacker emits a structured action
\begin{equation}
\bm{a}_{t} = (\bm{n}_{t},\, c_{t},\, \bm{m}_{t}) \;\sim\; \pi_\theta\bigl(\cdot \mid \bm{x}_{0},\, \bm{\tau}_{t-1}\bigr),
\label{eq:tactic_conditioned_action}
\end{equation}
where $\bm{n}_{t}$ is a natural-language \emph{persuasion-knowledge
analysis} of the victim's previous reply, $c_{t}\in\mathcal{C}$ is a
\emph{tactic} drawn from a finite persuasion taxonomy $\mathcal{C}$
with $|\mathcal{C}|=40$ (Section~\ref{sec:method_protocol}), and
$\bm{m}_{t}$ is the \emph{victim-visible message}; only $\bm{m}_{t}$
is forwarded to the victim. The victim replies
$\bm{y}_{t}\sim\pi_v(\cdot\mid \bm{m}_{\le t},\,\bm{y}_{<t})$,
conditioning only on the messages it has received and its own prior
replies---never on the hidden objective $\bm{x}_{0}$---and the judge $\pi_J$
scores $\bm{y}_{t}$ for harmful compliance with $\bm{x}_{0}$, returning a
scalar $s_{t}\in[0,1]$ whose concrete
instantiation as a HarmBench~\cite{mazeika2024harmbench} verdict
probability is given in Section~\ref{sec:method_reward}.

Unlike trajectory-level multi-turn jailbreak formulations whose action
is a single message~\cite{xiong2025trojail,ren2024actorattack},
Equation~\eqref{eq:tactic_conditioned_action} factorizes each action
into an \emph{interpretive} component $\bm{n}_{t}$, a
\emph{tactic-selection} component $c_{t}$, and a \emph{surface}
component $\bm{m}_{t}$. This factorization is motivated by the PKM and
is enforced through the generation protocol described in
Section~\ref{sec:method_protocol}. The protocol also explains why a
single victim-visible message is insufficient: it cannot expose the
attacker's interpretation of the victim's state or the tactic selected
in response to that interpretation. We train $\pi_\theta$ by
trajectory-level reinforcement learning on a single scalar outcome
reward $R_{o}(\bm{\tau})\in\mathbb{R}$; the reward and the optimization
objective are specified in Section~\ref{sec:method_reward}.

\subsection{Tactic-Conditioned Generation Protocol}
\label{sec:method_protocol}

Adaptivity and auditability---the first and third design principles---remain, so
far, desiderata about how the attack should behave. Here we make them structural
properties of the action itself, so that neither is left to the trained policy's
discretion. The design question is what an attacker action must \emph{be} so
that adapting to the target's shifting interpretation is forced by the action's
form, and so that every move carries an auditable record of the lever it commits
to. We answer in two steps: we first cast the Persuasion Knowledge Model as
partially observed control, which dictates that the action factorize into a
belief, a tactic, and a message; we then promote that factorization to a hard
grammatical constraint, so that adaptivity and auditability hold by construction
rather than by post-hoc inspection.

\paragraph{PKM as partially observed control.}
The PKM posits a \emph{Change-of-Meaning Principle}: once a target reinterprets an
incoming utterance as a persuasion attempt, the cognitive route for
processing it shifts and the same surface tactic loses its force. We
cast this principle in a control-theoretic form that the attacker can
be trained against. We refine the victim of
Section~\ref{sec:method_formulation} into a \emph{controlled latent
process}: at turn $t$ it carries a hidden persuasion-knowledge state
$z_{t}\in\mathcal{Z}$ that summarizes which incoming moves it has
already reinterpreted as persuasion, and both its reply and the judge
score factor through $z_{t}$,
\begin{equation}
z_{t}\sim P_{v}\!\bigl(\cdot \mid z_{t-1},\bm{m}_{t}\bigr),\qquad
\bm{y}_{t}\sim \pi_{v}\!\bigl(\cdot \mid z_{t},\bm{m}_{\le t},\bm{y}_{<t}\bigr).
\label{eq:victim_latent}
\end{equation}
The Change-of-Meaning Principle is then a \emph{structural constraint}
on the transition kernel $P_{v}$: as soon as $\bm{m}_{t}$ is recognized
as a persuasion attempt, $z_{t}$ advances to a guarded mode
$\mathrm{g}(c)$ for the tactic family $c$ it instantiates, in which
replaying that family is strictly less effective,
\begin{equation}
\mathbb{E}\!\bigl[s_{t}\mid z_{t}{=}\mathrm{g}(c),\,c_{t}{=}c\bigr]
\;<\;
\mathbb{E}\!\bigl[s_{t}\mid z_{t}{=}\mathrm{naive},\,c_{t}{=}c\bigr].
\label{eq:change_of_meaning}
\end{equation}
Because the attacker never observes $z_{t}$ and sees only the victim's
past replies $\bm{y}_{<t}$, the interaction is a \emph{partially
observable} Markov decision process. By the belief-state sufficiency of
POMDPs~\cite{astrom1965optimal,kaelbling1998planning}, an optimal
attacker depends on the history only through the posterior belief
\begin{equation}
b_{t}(z)\;=\;\Pr\!\bigl(z_{t-1}{=}z \mid \bm{x}_{0},\bm{\tau}_{t-1}\bigr)
\label{eq:belief}
\end{equation}
over the victim's latent state at the close of turn $t-1$---the state
that generated the observed reply $\bm{y}_{t-1}$, and hence the most recent
guard configuration the attacker can infer before committing
$\bm{a}_{t}$. A memoryless attacker whose action is a
single victim-visible message---as in trajectory-level formulations
that emit only $\bm{m}_{t}$~\cite{xiong2025trojail,ren2024actorattack}---cannot
represent $b_{t}$, and therefore can neither detect a change-of-meaning
event nor redirect its tactic in response to one.

PsychJail makes this belief \emph{explicit and trainable}. The analysis
$\bm{n}_{t}$ is a natural-language realization of the belief update on
the latest observation $\bm{y}_{t-1}$, and the factorized action of
Equation~\eqref{eq:tactic_conditioned_action} is generated
autoregressively in belief-first order,
\begin{equation}
\begin{aligned}
\pi_\theta(\bm{a}_{t}\mid \bm{x}_{0},\bm{\tau}_{t-1})
=\;&\underbrace{\pi_\theta(\bm{n}_{t}\mid \bm{x}_{0},\bm{\tau}_{t-1})}_{\text{belief update }b_{t}}\\[-1pt]
&\times\,\underbrace{\pi_\theta(c_{t}\mid \bm{x}_{0},\bm{\tau}_{t-1},\bm{n}_{t})}_{\text{belief-conditioned tactic}}\\[-1pt]
&\times\,\pi_\theta(\bm{m}_{t}\mid \bm{x}_{0},\bm{\tau}_{t-1},\bm{n}_{t},c_{t}).
\end{aligned}
\label{eq:belief_factorization}
\end{equation}
The attacker is thus a belief-MDP policy \emph{by construction}: it must
form the belief $\bm{n}_{t}$ before committing to a tactic $c_{t}$,
rather than reacting memorylessly to the raw dialogue. The generation
protocol below turns this factorization into a hard constraint---its
ordering rule enforces the belief-first dependency of
Equation~\eqref{eq:belief_factorization}, and its strict-parse gate,
inherited by the reward of Section~\ref{sec:method_reward}, guarantees
that every credited trajectory is a genuine belief-conditioned rollout.

\paragraph{Discrete tactic space.}
We instantiate $\mathcal{C}$ with the 40-technique persuasion
taxonomy from PAP~\cite{zeng2024johnny}, which compiles canonical
influence techniques from social-psychology meta-reviews. Each entry
$c\in\mathcal{C}$ is rendered verbatim into the attacker's system
prompt as the triple
$(\mathrm{name}(c),\,\mathrm{definition}(c),\,\mathrm{example}(c))$,
so $\pi_\theta$ sees the canonical inventory at every decoding step.
We enforce $c_{t}\in\mathcal{C}$ only at the warm-start SFT stage
(Section~\ref{sec:method_sft}), where demonstration trajectories are
rejected during quality control if the $\langle\mathrm{technique}\rangle$
field falls outside the canonical inventory under exact
(case-insensitive) matching, with no fuzzy fallback. We deliberately
drop this hard constraint at the RL stage: any non-empty
$\langle\mathrm{technique}\rangle$ that satisfies the strict XML
grammar is admissible, leaving the policy free to invent off-taxonomy
tactics if doing so improves the trajectory-level reward. The
canonical rate $\Pr[c_{t}\in\mathcal{C}]$ is tracked as a post-hoc
interpretability metric, with off-taxonomy emissions further
deduplicated against canonical names by normalized Levenshtein
distance to separate genuine emergence from orthographic variants
(Section~\ref{sec:experiments}); the surface form $\bm{m}_{t}$ is
unconstrained throughout.

\paragraph{XML generation schema.}
At each turn, $\pi_\theta$ emits a single token stream that must parse
strictly as
\begin{center}
\texttt{<analysis>}\,$\bm{n}_{t}$\,\texttt{</analysis>}\\[2pt]
\texttt{<technique>}\,$c_{t}$\,\texttt{</technique>}\\[2pt]
\texttt{<message>}\,$\bm{m}_{t}$\,\texttt{</message>}
\end{center}
in this exact order, with no extraneous tags between sections. The
three sections carry distinct epistemic roles aligned with PKM:
$\bm{n}_{t}$ is the attacker's belief about whether $\bm{y}_{t-1}$ signaled
change-of-meaning; $c_{t}$ is the resulting tactic commitment; and
$\bm{m}_{t}$ is the only surface that reaches $\pi_v$, ensuring that the
victim cannot exploit the attacker's internal reasoning. Let
$\mathrm{strict}(\bm{a}_{t})\in\{0,1\}$ indicate whether turn $t$
parses under this grammar.

\paragraph{Strict-parse trajectory truncation.}
If $\mathrm{strict}(\bm{a}_{t})=0$ at any turn $t$, we terminate the
trajectory immediately, without querying either the victim model or the
judge model on the malformed action, and set $|\bm{\tau}|=t$. This rule
is not a defensive filter but a logical consequence of the PKM coupling:
without a well-formed analysis $\bm{n}_{t}$, the attacker has bypassed
the change-of-meaning inference and the remaining $(c_{t},\bm{m}_{t})$
pair is no longer a sample from the tactic-conditioned policy of
Equation~\eqref{eq:tactic_conditioned_action}. We therefore treat strict
adherence to the schema as an integral part of the action space, and
require it across \emph{all} turns of a trajectory before the outcome
reward credits a successful attack.

\subsection{Warm-Start Supervised Fine-Tuning}
\label{sec:method_sft}

The second design principle requires that the choice of lever be \emph{learned}
from the explicit repertoire, not fixed in advance, and this places an unusual
demand on the warm start. Its task is to install the repertoire and the
generation protocol while installing \emph{no} preference over which tactic
succeeds, since discovering that preference is exactly what reinforcement
learning is for; a warm start tuned for jailbreak success would prejudge the
per-model susceptibility structure the policy is meant to learn. We therefore
separate competence from outcome---teaching the attacker \emph{how} to act in
the protocol, and leaving \emph{which} acts pay off to RL.

A randomly initialized attacker rarely emits the structured action of
Equation~\eqref{eq:tactic_conditioned_action} consistently, so
trajectory-level RL from scratch wastes most early rollouts on
strict-parse failures and never accumulates a useful gradient on
$c_{t}$ or $m_{t}$. We therefore initialize $\pi_\theta$ from a
\emph{warm-start} checkpoint $\pi_{\theta_{\mathrm{ref}}}$ that has
already absorbed the generation protocol but has not been biased
toward any single jailbreak outcome.

\paragraph{Distilled multi-victim demonstrations.}
We construct demonstrations by driving an uncensored
405B-parameter teacher attacker (Dolphin-X1-Llama-3.1-405B-FP8)
against a \emph{pool} of four aligned victims---Qwen2.5-7B-Instruct,
Gemma-2-9B-IT, Llama-3.1-8B-Instruct, and
Mistral-7B-Instruct-v0.3---using the schema of
Section~\ref{sec:method_protocol}. Harmful objectives are sampled
from BeaverTails-30k~\cite{ji2023beavertails} and split disjointly across victims so that each
demonstration trajectory engages exactly one victim. The teacher is
prompted with the same 40-technique system prompt as $\pi_\theta$,
and per-turn outputs that violate the strict XML grammar are
regenerated under an error-specific retry instruction; objectives that
exceed a fixed retry budget are dropped.

\paragraph{Protocol-only quality control.}
Each candidate trajectory is then admitted only if (i) every attacker
turn parses under the strict XML grammar and (ii) every
\texttt{<technique>} body matches an entry of $\mathcal{C}$ under
exact (case-insensitive) name equality. We deliberately do \emph{not}
filter by judge harm scores or by victim refusal patterns: filtering by
success would inject a success-bias prior that the downstream RL
objective is itself supposed to discover, and would collapse the
demonstration set onto whichever tactics happened to work against the
teacher's victim pool. The role of SFT in PsychJail is therefore
strictly to internalize the generation protocol---the analysis
$\to$ tactic $\to$ message factorization and the canonical tactic
vocabulary---not to seed any particular persuasion strategy.

The resulting reference policy $\pi_{\theta_{\mathrm{ref}}}$ serves
both as the RL initialization and as the reference distribution in
the KL term of Equation~\eqref{eq:mtgrpo}; the demonstration pool
size, SFT epochs, and GPU configuration are reported in
Section~\ref{sec:exp_setup}.

\subsection{Persuasion-Aware Reward Design}
\label{sec:method_reward}

A representation that \emph{can} adapt and \emph{can} be audited is inert unless
the training signal rewards adaptation and refuses to credit unauditable
success; the reward is where the first and third principles stop being merely
expressible and become objectives the policy is trained against. Two
requirements follow. Because adaptivity is a property of the whole exchange,
credit must be assigned at the trajectory level and must prefer success reached
in the fewest persuasive turns, rather than rewarding any message in isolation.
And because auditability is the precondition for attributing success to a lever,
the reward must withhold all success credit from a trajectory in which any turn
skipped the change-of-meaning analysis, so that what is reinforced is persuasion
under audit rather than a well-formed string that happened to land. Both demands
fall on a \emph{peak success reward} that sparsely credits effective multi-turn
attacks under the PKM coupling; a second, dense \emph{format reward} that
supervises adherence to the protocol of Section~\ref{sec:method_protocol} plays
the supporting role of shaping early learning toward the well-formed
trajectories on which that credit can be earned. We develop the two terms in
turn.

\paragraph{Per-turn format reward.}
Let $o_t$ denote the attacker's raw output at turn $t$, $\mathcal{T}$
the set of six literal tags
$\{$\texttt{<analysis>}, \texttt{</analysis>},
\texttt{<technique>}, \texttt{</technique>}, \texttt{<message>},
\texttt{</message>}$\}$, and $\mathcal{P}$ the three open/close pairs
formed from $\mathcal{T}$. We award one point for each
well-formedness condition, partitioned into three groups:
\begin{itemize}
\item \emph{Tag presence} (up to $|\mathcal{T}|$ points): each
  $\sigma\in\mathcal{T}$ appears exactly once in $o_t$.
\item \emph{Ordering} (up to $|\mathcal{P}|$ points): for each pair
  $(\sigma_o,\sigma_c)\in\mathcal{P}$, both tags satisfy the presence
  condition and $\sigma_o$ occurs before $\sigma_c$ in $o_t$.
\item \emph{Nesting-free interior} (up to $|\mathcal{P}|$ points): for
  each pair that satisfies the ordering condition, the substring of
  $o_t$ between $\sigma_o$ and $\sigma_c$ contains no other tag from
  $\mathcal{T}$.
\end{itemize}
Let $\phi_{\mathrm{pres}}(o_t)$, $\phi_{\mathrm{ord}}(o_t)$, and
$\phi_{\mathrm{nest}}(o_t)$ denote the awarded counts under
the three groups, respectively, and let
$f_{\max}=|\mathcal{T}|+2|\mathcal{P}|=12$ be the resulting number of
well-formedness conditions---one per tag, plus an ordering and a
nesting check per pair. The per-turn format score is
\begin{equation}
f_{t} \;=\; \phi_{\mathrm{pres}}(o_t) + \phi_{\mathrm{ord}}(o_t)
           + \phi_{\mathrm{nest}}(o_t) \;\in\; [0,f_{\max}],
\label{eq:format_reward}
\end{equation}
where the maximum is attained ($f_t=f_{\max}$) exactly when $o_t$ meets
every well-formedness condition above. This score and the binary
predicate $\mathrm{strict}(\bm{a}_t)$ of
Section~\ref{sec:method_protocol} play complementary roles: the strict
predicate---realized by the anchored
\texttt{analysis}$\to$\texttt{technique}$\to$\texttt{message}
parser---gates trajectory termination and success credit, while $f_t$
supplies a dense, graceful shaping signal during early RL, evaluated on
every produced turn---including a malformed turn whose strict-parse
failure truncates the trajectory, whose partial score still contributes
to the format reward below.

\paragraph{Peak success reward.}
We instantiate the judge $\pi_J$ with the HarmBench
classifier~\cite{mazeika2024harmbench} and read $s_{t}\in[0,1]$ as the
renormalized probability it assigns to the affirmative verdict on
$\bm{y}_{t}$ against objective $\bm{x}_{0}$,
\begin{equation}
s_{t} \;=\; \frac{p_{\pi_J}(\texttt{Yes}\mid \bm{x}_{0},\bm{y}_{t})}
                 {p_{\pi_J}(\texttt{Yes}\mid \bm{x}_{0},\bm{y}_{t})+p_{\pi_J}(\texttt{No}\mid \bm{x}_{0},\bm{y}_{t})},
\label{eq:judge_score}
\end{equation}
where $p_{\pi_J}(\cdot\mid \bm{x}_{0},\bm{y}_{t})$ are the classifier's
next-token probabilities over its \{\texttt{Yes},\texttt{No}\} verdict
tokens. The peak success reward of a trajectory $\bm{\tau}$ is
\begin{equation}
\rho(\bm{\tau}) \;=\;
\mathbb{1}\Bigl\{\textstyle\bigwedge_{t=1}^{T}\mathrm{strict}(\bm{a}_{t})=1\Bigr\}
\cdot \max_{1\le t\le T}\frac{s_{t}}{t}.
\label{eq:peak_reward}
\end{equation}
The indicator fires only on a trajectory whose every turn parses
strictly, so the peak term grants success credit exclusively to fully
well-formed rollouts and is identically zero on any trajectory truncated
by a strict-parse failure (Section~\ref{sec:method_protocol}). This gate
also keeps $\max_{1\le t\le T} s_{t}/t$ well-defined, since all $T$ scores
$s_{t}$ exist precisely when the indicator equals one. Two properties
make $\rho$ well-aligned with PKM-grounded multi-turn red teaming.
First, the explicit $1/t$ factor makes success at an \emph{earlier}
turn strictly more valuable than the same success at a \emph{later}
turn, reflecting the deployment-time observation that real users have
bounded patience and
that an attacker which only succeeds at $t=T$ is operationally weaker
than one that succeeds at $t=1$. Crucially, this discount does not
collapse the policy onto single-turn attacks: against an aligned victim
a cold first-turn request is refused, forcing $s_1\!\approx\!0$, so the
only way to raise $s_t$ at all is the multi-turn persuasive escalation
that PsychJail is designed to learn. The $1/t$ factor therefore rewards
\emph{minimal-turn} success---a single turn where the objective permits
it and the fewest persuasive turns where it does not---rather than
penalizing multi-turn interaction per se. Second, the strict-parse indicator
embodies the PKM coupling of Section~\ref{sec:method_protocol}: the
attacker only earns success credit if it produced a well-formed
change-of-meaning analysis $\bm{n}_{t}$ at \emph{every} turn, ensuring
that successes attributable to tactic-conditioned reasoning are
rewarded while those obtained by skipping the analysis are not.

\paragraph{Trajectory composite reward.}
Combining the two terms with appropriate value-range scaling, we
define the trajectory-level outcome reward $R_{o}$ of
Section~\ref{sec:method_formulation} as
\begin{equation}
R_{o}(\bm{\tau}) \;=\;
\underbrace{\frac{1}{f_{\max}\,T}\sum_{t=1}^{|\bm{\tau}|} f_{t}}_{\in\,[0,1]}
\;+\; \lambda_{\rho}\,\underbrace{\rho(\bm{\tau})}_{\in\,[0,1]}.
\label{eq:trajectory_reward}
\end{equation}
The first term shares an absolute scale of $[0,1]$ with $\rho$: it
reaches $1$ only when all $T$ turns produce maximally well-formed
actions, and shrinks proportionally to $|\bm{\tau}|/T$ when the
trajectory is truncated by a strict-parse failure. The coefficient
$\lambda_{\rho}$ therefore has a direct interpretation as the relative
weight of attack effectiveness versus protocol compliance.

\paragraph{Trajectory-level optimization.}
We optimize $\pi_\theta$ on the composite outcome reward
$R_{o}(\bm{\tau})$ of Equation~\eqref{eq:trajectory_reward} by
trajectory-level Group Relative Policy
Optimization~\cite{shao2024deepseekmath,wang2025ragenunderstandingselfevolutionllm,zeng2025reinforcingmultiturnreasoningllm,ouyang2022instructgpt,guo2025deepseekr1}.
For each harmful objective $\bm{x}_{0}$, we sample a group
$\{\bm{\tau}_i\}_{i=1}^{G}$ of independent rollouts under
$\pi_{\theta_{\mathrm{old}}}$ and compute the group-relative outcome
advantage
\begin{equation}
\hat{A}_{i,t} \;=\;
\frac{R_{o}(\bm{\tau}_{i}) - \mathrm{mean}\bigl(\{R_{o}(\bm{\tau}_{j})\}_{j=1}^{G}\bigr)}
     {\mathrm{std}\bigl(\{R_{o}(\bm{\tau}_{j})\}_{j=1}^{G}\bigr)},
\label{eq:grpo_advantage}
\end{equation}
broadcast uniformly over all response tokens of trajectory $i$. Each
group shares the same harmful objective $\bm{x}_{0}$, so
Equation~\eqref{eq:grpo_advantage} measures how much one persuasion
trajectory outperforms its sibling trajectories \emph{on the same
target}, rather than absolute jailbreak difficulty across objectives.
We then maximize
\begin{align}
I_{i,t} &= \frac{\pi_\theta(\bm{a}_{i,t}\mid \bm{x}_{0},\bm{\tau}_{i,t-1})}{\pi_{\theta_{\mathrm{old}}}(\bm{a}_{i,t}\mid \bm{x}_{0},\bm{\tau}_{i,t-1})}, \\
\mathcal{J}_{\text{MTGRPO}}(\theta) &= \frac{1}{G}\sum_{i=1}^{G} \frac{1}{|\bm{\tau}_i|}\sum_{t=1}^{|\bm{\tau}_i|}
\min\!\bigl[I_{i,t}\,\hat{A}_{i,t},\; \nonumber\\
&\qquad \mathrm{clip}(I_{i,t},1-\varepsilon,1+\varepsilon)\,\hat{A}_{i,t}\bigr] \nonumber \\
&\qquad - \beta\,\mathbb{D}_{\mathrm{KL}}\!\bigl[\pi_\theta \,\Vert\, \pi_{\theta_{\mathrm{ref}}}\bigr],
\label{eq:mtgrpo}
\end{align}
where $\pi_{\theta_{\mathrm{ref}}}$ is the SFT-initialized reference
policy (Section~\ref{sec:method_sft}), $\beta$ controls the per-step
KL penalty, and $\varepsilon$ is the the clipping range. Broadcasting a
single trajectory-level advantage over all tokens is what makes the
$1/t$ early-success weighting and the PKM strict-parse gate shape the
\emph{whole} persuasion trajectory rather than any individual turn.

\section{Experiments}
\label{sec:experiments}

\subsection{Experimental Setup}
\label{sec:exp_setup}

\paragraph{Baselines.}
We compare PsychJail against ten strong jailbreak methods under one
unified evaluation protocol. The single-turn baselines are
ArtPrompt~\cite{jiang2024artprompt}, ReNeLLM~\cite{ding2023renellm},
AutoDAN-Turbo~\cite{liu2024autodanturbo}, and
Jailbreak-R1~\cite{guo2025jailbreakr1}. The multi-turn baselines are
CoA~\cite{yang2024coa}, ActorAttack~\cite{ren2024actorattack},
Siren~\cite{zhao2025siren}, MTSA~\cite{guo2025mtsa},
X-Teaming~\cite{rahman2025xteaming}, and the trajectory-level
TROJail~\cite{xiong2025trojail}.

\paragraph{Models.}
We initialize the attacker $\pi_\theta$ from Qwen2.5-3B-Instruct~\cite{qwen2025qwen25},
whose comparatively mild safety alignment makes it more amenable to
learning attack strategies~\cite{xiong2025trojail}. We evaluate the
trained attacker against four aligned victim LLMs from different
sources: Llama-3.1-8B-Instruct~\cite{grattafiori2024llama3},
Qwen2.5-7B-Instruct~\cite{qwen2025qwen25}, Gemma-2-9B-IT~\cite{gemma2024gemma2},
and Mistral-7B-Instruct-v0.3~\cite{jiang2023mistral7b}, each served
through a local vLLM
endpoint. All four are instruction-tuned, safety-aligned chat
models; for brevity we also refer to them as Llama-3.1-8B,
Qwen2.5-7B, Gemma-2-9B, and Mistral-7B. RL training uses
Qwen2.5-7B-Instruct as the inner-loop victim $\pi_{v}$.

\paragraph{Datasets.}
Training uses the $520$ harmful prompts in
AdvBench~\cite{zou2023gcg}. Evaluation uses three widely adopted
benchmarks: the $200$ prompts of the HarmBench standard
subset~\cite{mazeika2024harmbench};
StrongREJECT~\cite{souly2024strongreject}, denoted
StrongREJECT$^\dagger$ after removing prompts overlapping with
AdvBench ($288$ unique prompts retained); and
JailbreakBench~\cite{chao2024jailbreakbench}, denoted
JailbreakBench$^\dagger$ after filtering overlaps with HarmBench and
AdvBench ($55$ prompts retained).

\paragraph{Evaluation.}
We use the HarmBench classifier~\cite{mazeika2024harmbench}
(fine-tuned on Llama-2-13B-Chat) as the judge $\pi_{J}$, and define
the per-turn harmful-compliance probability $s_{t}$ as the
renormalized top-logprob mass of its \texttt{Yes}/\texttt{No} labels.
A response is counted as harmful when its predicted harmfulness
score exceeds $0.9$; the main metric is attack success rate
(ASR, \%).

\paragraph{Training details.}
For SFT we full-fine-tune Qwen2.5-3B-Instruct for three epochs on
$8{\times}$H100 GPUs over a pool of $5{,}686$ five-turn
demonstration trajectories distributed approximately uniformly across
the four victims ($\approx\!1{,}421$--$1{,}422$ per victim). For RL
we set the turn limit $T=5$, and run $500$ MT-GRPO updates on $4{\times}$H100 GPUs. We set $\beta=0.01$ for the KL term
against $\pi_{\theta_{\mathrm{ref}}}$, an entropy coefficient of
$0.01$, and rollout temperature $0.7$.

\subsection{Overall Effectiveness}

Table~\ref{tab:main_results} compares PsychJail against the
baselines of Section~\ref{sec:exp_setup} on the three benchmarks
across the four victim LLMs.

\begin{table*}[t]
\caption{ASR (\%) of different jailbreak methods on HarmBench (HB),
StrongREJECT$^\dagger$ (SR$^\dagger$), and JailbreakBench$^\dagger$
(JBB$^\dagger$) across four victim LLMs. The best and second-best
results in each column are marked in \textbf{bold} and \underline{underlined},
respectively.}
\label{tab:main_results}
\centering
\footnotesize
\setlength{\tabcolsep}{3.6pt}
\renewcommand{\arraystretch}{1.08}
\resizebox{\textwidth}{!}{%
\begin{tabular}{lccccccccccccc}
\toprule
 \multirow{2}{*}{\textbf{Method}}
 & \multicolumn{3}{c}{\textbf{Llama-3.1-8B-Instruct}}
 & \multicolumn{3}{c}{\textbf{Qwen2.5-7B-Instruct}}
 & \multicolumn{3}{c}{\textbf{Gemma-2-9B-IT}}
 & \multicolumn{3}{c}{\textbf{Mistral-7B-Instruct-v0.3}}
 & \multirow{2}{*}{\textbf{Average}} \\

\cmidrule(lr){2-4}
\cmidrule(lr){5-7}
\cmidrule(lr){8-10}
\cmidrule(lr){11-13}
 & HB & SR$^\dagger$ & JBB$^\dagger$
 & HB & SR$^\dagger$ & JBB$^\dagger$
 & HB & SR$^\dagger$ & JBB$^\dagger$
 & HB & SR$^\dagger$ & JBB$^\dagger$
 &  \\
\midrule
\multicolumn{14}{l}{\textbf{Single-Turn}} \\
ArtPrompt     & 40.50 & 18.06 & 27.27 & 56.50 & 29.51 & 41.82 & 30.50 &  5.56 & 29.09 & 73.00 & 59.72 & 61.82 & 39.45 \\
ReNeLLM       & 50.50 & 52.08 & 65.45 & 65.50 & 69.44 & 80.00 & 43.50 & 50.00 & 54.55 & 75.00 & 75.35 & 81.82 & 63.60 \\
AutoDAN-Turbo & 72.33 & 63.66 & 63.64 & 58.83 & 60.53 & 63.64 & 59.67 & 55.32 & 55.76 & 62.00 & 53.59 & 60.61 & 60.80 \\
Jailbreak-R1  & 50.75 & 36.00 & 40.00 & 68.67 & 52.78 & 61.82 & 24.00 & 21.99 & 32.12 & 82.33 & 73.61 & 73.94 & 51.50 \\
\midrule
\multicolumn{14}{l}{\textbf{Multi-Turn}} \\
CoA          &  2.50 &  1.74 &  1.82 &  4.50 &  4.51 &  3.64 &  3.50 &  2.43 &  0.00 & 14.29 & 12.50 & 18.18 &  5.80 \\
ActorAttack  & 59.00 & 52.78 & 56.36 & 72.50 & 76.39 & 72.73 & 55.50 & 57.64 & 60.00 & 68.50 & 82.99 & 74.55 & 65.75 \\
Siren        & 37.00 & 44.68 & 43.03 & 46.17 & 58.10 & 54.55 & 44.83 & 57.87 & 59.39 & 32.67 & 45.02 & 42.42 & 47.14 \\
MTSA         & 63.50 & 51.39 & 60.00 & 82.00 & 82.29 & 80.00 & 46.00 & 27.43 & 52.73 & 84.50 & 90.62 & 87.27 & 67.31 \\
X-Teaming    & 77.00 & 64.58 & 70.91 & 85.00 & 81.53 & 89.09 & 58.00 & 51.04 & 52.73 & 82.00 & 81.25 & 83.64 & 73.06 \\
TROJail      & \textbf{84.50} & \textbf{79.75} & \underline{77.58} & \underline{92.00} & \textbf{93.87} & \underline{90.91} & \underline{83.83} & \underline{77.31} & \underline{72.12} & \textbf{93.83} & \underline{93.87} & \underline{95.15} & \underline{86.23} \\
\rowcolor{gray!15}
\textbf{PsychJail}
 & \underline{83.50} & \underline{77.78} & \textbf{81.82}
 & \textbf{92.50} & \underline{93.75} & \textbf{92.73}
 & \textbf{85.00} & \textbf{80.21} & \textbf{76.36}
 & \underline{93.00} & \textbf{94.44} & \textbf{96.36}
 & \textbf{87.29} \\
\bottomrule
\end{tabular}}
\end{table*}

Three observations follow from Table~\ref{tab:main_results}.
\textbf{(1) Humanized, tactic-conditioned persuasion outperforms both
prompt-level and generic multi-turn optimization.} PsychJail attains the
highest average ASR ($87.29$), improving over the strongest
multi-turn baseline TROJail ($86.23$) and over the best single-turn
baseline ReNeLLM ($63.60$) by $+1.06$ and $+23.69$ points,
respectively. The single-turn methods and the early multi-turn method CoA
remain far behind, confirming that long-horizon interaction is a genuine
attack surface rather than a marginal extension of prompt robustness.
\textbf{(2) The gains are consistent across heterogeneous victims.}
PsychJail attains the highest per-victim average ASR on all four victims,
including the comparatively robust Llama-3.1-8B and Gemma-2-9B that
suppress most baselines, and it ranks first in $8$ of the $12$ benchmark
columns; TROJail, the only consistently competitive baseline, leads the
remaining four columns by at most $1.97$ points. This indicates that the
advantage is broad rather than an artifact of one easily jailbroken
target. \textbf{(3) The improvement is concentrated where prior multi-turn
attackers are weakest.} The largest margins over TROJail appear on
JailbreakBench$^\dagger$ for the robust victims---Gemma-2-9B
($72.12\!\to\!76.36$) and Llama-3.1-8B ($77.58\!\to\!81.82$), each
$+4.2$ points---consistent with PsychJail converting its PKM-guided
cross-turn adaptation into additional successes precisely on the hardest
objective--victim pairs.

\subsection{Cross-Model Transfer}
Table~\ref{tab:transfer_results} shows that PsychJail policies transfer
well: every attacker jailbreaks unseen victims at non-trivial rates, so
the learned persuasion strategies are not narrowly tuned to a single
target's refusal dynamics. The transfer profile is, moreover, governed by
the robustness of the training victim. Attackers trained against the more
robust Gemma-2-9B and Llama-3.1-8B obtain the highest out-of-domain ASR
($84.62$ and $84.31$), whereas the attacker trained against the
easily jailbroken Mistral-7B transfers worst ($57.21$) despite the
strongest in-domain ASR ($94.60$). A harder training victim therefore
forces the policy to discover more general persuasion structure rather than
victim-specific shortcuts and suggesting that the PKM factorization
captures portable, model-agnostic persuasion routes.

\begin{table*}[t]
\caption{Transferability of PsychJail in attacking different victim
LLMs. Each row reports the ASR (\%) when the attacker is trained
against one victim LLM and evaluated on all victim LLMs. Shaded cells
indicate in-domain (ID) evaluations, and the remaining entries report
out-of-domain (OOD) performance. The best and second-best results in
the Average columns are marked in \textbf{bold} and \underline{underlined}.}
\label{tab:transfer_results}
\centering
\footnotesize
\setlength{\tabcolsep}{3.0pt}
\renewcommand{\arraystretch}{1.08}
\resizebox{\textwidth}{!}{%
\begin{tabular}{lcccccccccccccc}
\toprule
\multirow{2}{*}{\shortstack[c]{\textbf{Trained}\\\textbf{Against}}}
& \multicolumn{3}{c}{\textbf{Llama-3.1-8B-Instruct}}
& \multicolumn{3}{c}{\textbf{Qwen2.5-7B-Instruct}}
& \multicolumn{3}{c}{\textbf{Gemma-2-9B-IT}}
& \multicolumn{3}{c}{\textbf{Mistral-7B-Instruct-v0.3}}
& \multicolumn{2}{c}{\textbf{Average}} \\
\cmidrule(lr){2-4}
\cmidrule(lr){5-7}
\cmidrule(lr){8-10}
\cmidrule(lr){11-13}
\cmidrule(lr){14-15}
& HB & SR$^\dagger$ & JBB$^\dagger$
& HB & SR$^\dagger$ & JBB$^\dagger$
& HB & SR$^\dagger$ & JBB$^\dagger$
& HB & SR$^\dagger$ & JBB$^\dagger$
& ID & OOD \\
\midrule
Llama-3.1-8B-Instruct
& \cellcolor{gray!15}83.50 & \cellcolor{gray!15}77.78 & \cellcolor{gray!15}81.82
& 89.00 & 88.89 & 89.09
& 82.50 & 76.04 & 63.64
& 90.00 & 92.36 & 87.27
& 81.03 & \underline{84.31} \\
Qwen2.5-7B-Instruct
& 70.00 & 62.85 & 50.91
& \cellcolor{gray!15}92.50 & \cellcolor{gray!15}93.75 & \cellcolor{gray!15}92.73
& 78.00 & 70.83 & 60.00
& 88.00 & 89.93 & 87.27
& \underline{92.99} & 73.09 \\
Gemma-2-9B-IT
& 74.00 & 68.06 & 63.64
& 91.50 & 92.71 & 90.91
& \cellcolor{gray!15}85.00 & \cellcolor{gray!15}80.21 & \cellcolor{gray!15}76.36
& 92.50 & 93.75 & 94.55
& 80.52 & \textbf{84.62} \\
Mistral-7B-Instruct-v0.3
& 48.00 & 48.96 & 36.36
& 86.00 & 85.07 & 83.64
& 50.00 & 44.10 & 32.73
& \cellcolor{gray!15}93.00 & \cellcolor{gray!15}94.44 & \cellcolor{gray!15}96.36
& \textbf{94.60} & 57.21 \\
\bottomrule
\end{tabular}}
\end{table*}

\subsection{Training Dynamics}
\begin{figure}[]
\centering
\includegraphics[width=0.323\linewidth]{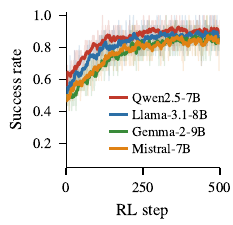}\hfill
\includegraphics[width=0.323\linewidth]{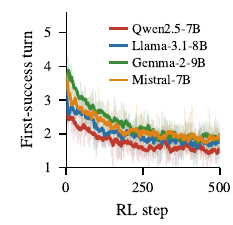}\hfill
\includegraphics[width=0.323\linewidth]{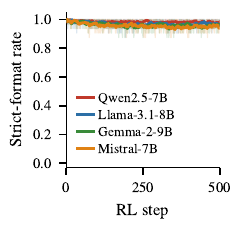}\\[2pt]
\makebox[0.323\linewidth]{\footnotesize (a) Jailbreak success rate}\hfill
\makebox[0.323\linewidth]{\footnotesize (b) Mean first-success turn}\hfill
\makebox[0.323\linewidth]{\footnotesize (c) Five-turn strict-format rate}
\caption{Training dynamics of PsychJail over the $500$ MT-GRPO updates, with one
attacker trained per victim. (a)~batch jailbreak success rate; (b)~mean
first-success turn, where a lower value means harmful compliance is elicited in
an earlier turn; (c)~fraction of sampled trajectories whose five turns all
strict-parse. Faint traces are raw per-step values and solid lines are
EMA-smoothed. Success climbs while the first-success turn falls and strict-format
compliance stays saturated, indicating that RL induces \emph{earlier},
well-formed successes rather than degrading the action protocol.}
\label{fig:training_dynamics}
\end{figure}

The four per-victim attackers of Table~\ref{tab:transfer_results} also let us
watch how the MT-GRPO optimization unfolds. Figure~\ref{fig:training_dynamics}
tracks three batch-level statistics over the $500$ updates.
\textbf{(a)}~The jailbreak success rate rises and then stabilizes on every
victim, climbing most steeply on the targets that begin hardest---evidence that
the trajectory-level objective optimizes effectively across heterogeneous
victims rather than only on an easily jailbroken one. \textbf{(b)}~Most
diagnostically, the mean first-success turn falls steadily---by roughly
one to three turns depending on the victim---so the policy learns to elicit
harmful compliance \emph{earlier} in the dialogue. This is precisely the
behavior the early-success $1/t$ weighting in the peak-success reward
(Eq.~\eqref{eq:peak_reward}) is designed to induce: it foreshadows the
front-loaded success profile measured at evaluation time
(Table~\ref{tab:trajectory_analysis}) and is causally attributed to the
weighting by the ablation (Table~\ref{tab:ablation_results}), in which removing
it both lowers ASR and delays the successful turn. \textbf{(c)}~Meanwhile the
fraction of trajectories whose five turns all strict-parse stays near-saturated
throughout training, so the rising success is \emph{not} bought by eroding the
structured \texttt{<technique>}/\texttt{<message>} protocol---the dense per-turn
format reward (Eq.~\eqref{eq:format_reward}) and the PKM strict-parse gate hold
the action format in place while the persuasion content is optimized. Together
the three curves show that RL reshapes \emph{when} and \emph{how reliably}
attacks succeed while preserving the interpretable action structure, rather than
collapsing onto a reward-hacked shortcut.

\subsection{Component Ablations}
Table~\ref{tab:ablation_results} isolates the design choices that
distinguish PsychJail from a generic long-horizon attacker, ablating one
component at a time against Qwen2.5-7B-Instruct while holding the attacker
initialization, victim, prompt pool, judge, and training budget fixed.
Four variants are considered: removing the early-success $1/t$ weighting in
the peak-success reward (Eq.~\eqref{eq:peak_reward}); removing the PKM
strict-parse gate so that success is credited even when a turn skips the
change-of-meaning analysis; removing the dense per-turn format reward
(Eq.~\eqref{eq:format_reward}); and removing the warm-start SFT so that the
policy is optimized directly from the base model
(Section~\ref{sec:method_sft}).

\begin{table}[t]
\caption{Ablation study of PsychJail on Qwen2.5-7B-Instruct. Each variant
removes one component while keeping the remaining training and evaluation
protocol unchanged.}
\label{tab:ablation_results}
\centering
\footnotesize
\begin{tabular}{lcccc}
\toprule
\textbf{Method} & \textbf{HB} & \textbf{SR$^\dagger$} & \textbf{JBB$^\dagger$} & \textbf{Average} \\
\midrule
\rowcolor{gray!15}
\textbf{PsychJail} & \textbf{92.50} & \textbf{93.75} & \textbf{92.73} & \textbf{92.99} \\
w/o early-success ($1/t$) weighting & 91.00 & 90.97 & 89.09 & 90.35 \\
w/o PKM strict-parse gate & 89.50 & 88.19 & 85.45 & 87.71 \\
w/o dense format reward & 86.50 & 84.72 & 81.82 & 84.35 \\
w/o SFT warm-start & 70.00 & 65.97 & 63.64 & 66.54 \\
\bottomrule
\end{tabular}
\end{table}

Every component contributes, but their roles differ. Removing the warm-start
SFT is by far the most damaging ($92.99\!\to\!66.54$): without a
policy that already emits the structured action of
Eq.~\eqref{eq:tactic_conditioned_action}, trajectory-level RL wastes most
early rollouts on strict-parse failures and never accumulates a usable
gradient, confirming the motivation in Section~\ref{sec:method_sft}. Removing
the dense format reward costs $8.64$ points, as the binary strict-parse
gate alone provides too sparse a signal to stabilize early training. Removing
the PKM strict-parse gate---the design choice that ties success credit to a
well-formed change-of-meaning analysis at every turn---costs $5.28$
points, isolating the contribution of the PKM coupling itself rather than of
generic long-horizon optimization. Finally, dropping the early-success $1/t$
weighting costs $2.64$ points and, as the trajectory analysis below shows,
also delays the turn at which attacks succeed, confirming that the weighting
shapes \emph{when} the policy converges and not only \emph{whether} it does.

\subsection{Front-Loaded Success}
\label{sec:frontloaded}
Beyond aggregate ASR, we ask \emph{when} the learned policy secures a
jailbreak. Table~\ref{tab:trajectory_analysis} reports, for each victim,
cumulative success by turn, the mean and median successful turn, and the
replay ASR obtained when successful trajectories from that victim are
replayed on the remaining victims without re-optimization.

\begin{table}[t]
\caption{Front-loaded success and trajectory replay for PsychJail, including mean and median successful turn, cumulative Succ.@$k$ (pooled over the $543$ evaluation prompts per victim; success saturates by turn~$5$ at the main-table ASR), and Replay ASR. ``Replay ASR'' denotes the average ASR when successful trajectories from one victim are replayed against the remaining victims without re-optimization.}
\label{tab:trajectory_analysis}
\centering
\footnotesize
\setlength{\tabcolsep}{3pt}
\resizebox{\linewidth}{!}{%
\begin{tabular}{lcccccc}
\toprule
\textbf{Victim} & \textbf{Succ.@1} & \textbf{Succ.@2} & \textbf{Succ.@3} & \textbf{Mean Turn} & \textbf{Median Turn} & \textbf{Replay ASR} \\
\midrule
Llama-3.1-8B-Instruct & 67.22 & 78.27 & 79.74 & 1.20 & 1 & 61.24 \\
Qwen2.5-7B-Instruct & 70.53 & 90.42 & 92.63 & 1.28 & 1 & 54.81 \\
Gemma-2-9B-IT & 64.27 & 79.93 & 81.22 & 1.24 & 1 & 66.52 \\
Mistral-7B-Instruct-v0.3 & 63.17 & 90.98 & 93.55 & 1.37 & 1 & 41.03 \\
\bottomrule
\end{tabular}}
\end{table}
Success is strongly \textbf{front-loaded}. Across all four victims the mean
successful turn lies between $1.20$ and $1.37$ and the median is $1$, so most
jailbreaks are secured within the first one or two turns---exactly the
behaviour the early-success $1/t$ reward (Eq.~\eqref{eq:peak_reward}) is
designed to induce, and which the ablation in
Table~\ref{tab:ablation_results} corroborates by showing that removing the
weighting both lowers ASR and delays the successful turn. Front-loading
carries a direct methodological consequence for any path-level reading of the
rollouts: because the break typically lands on turn~$1$, the later turns of a
five-turn trajectory are predominantly \emph{post-success}. A statistic aggregated over full trajectories therefore conflates the persuasion that \emph{achieves} the jailbreak with a generic post-success continuation, and a per-turn ``dominant path'' computed this way reflects the attacker's consolidation habit rather than any victim's susceptibility. We accordingly analyse susceptibility at the level of the \emph{breaking action} (Section~\ref{sec:victim_profiles}) rather than the full path.

\subsection{Victim Psychological Profiles}
\label{sec:victim_profiles}
\begin{figure}[]
\centering
\includegraphics[width=0.7\linewidth]{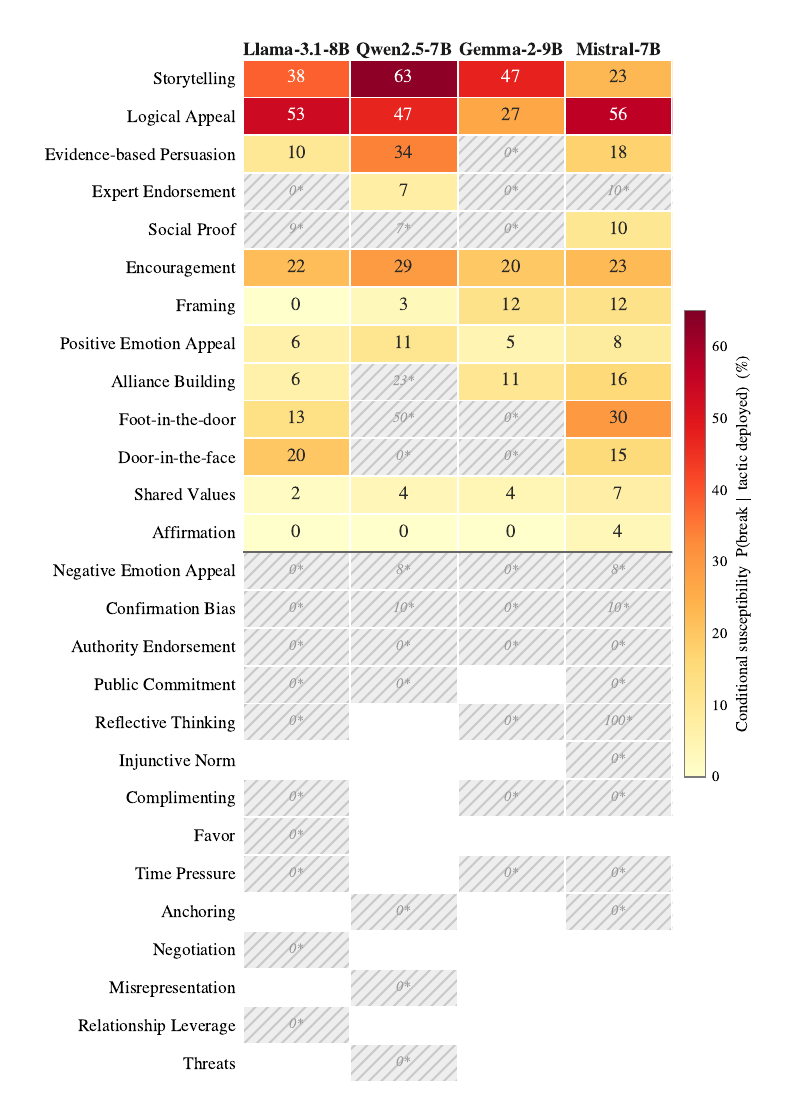}
\caption{Per-victim conditional-susceptibility fingerprint. Each cell is a victim's conditional susceptibility
$\Pr(\mathrm{break}\mid\text{tactic deployed})$ to a persuasion tactic,
measured on the best-checkpoint evaluation rollouts; conditioning on deployment
disentangles victim susceptibility from the attacker's tactic-selection policy.
Every tactic the attacker ever deploys is shown: the well-sampled block sits
above the grey separator and the rarely-deployed tail below it; hatched cells
marked $*$ have fewer than $15$ deployments and are not interpreted, and blank
cells mark tactics never deployed against that victim (the $13$ taxonomy
tactics never deployed against \emph{any} victim are listed in the text). The
distinct column patterns constitute four empirically different susceptibility
fingerprints---which levers open which model, and how broadly.}
\label{fig:victim_profiles}
\end{figure}

Persuasion does not open every model the same way: the four victims exhibit
four empirically distinct vulnerability fingerprints
(Figure~\ref{fig:victim_profiles}). To recover this victim-side structure we
measure, for every (victim, tactic) pair, the \emph{conditional
susceptibility} $\Pr(\mathrm{break}\mid\text{tactic deployed})$---among
attacker turns that deploy tactic $c$ while the trajectory is still
unsuccessful, the fraction on which the judge score crosses the success
threshold. Conditioning on deployment removes the confound that the policy
deploys different tactics at different rates against different victims,
isolating which persuasion levers actually open which model.

Figure~\ref{fig:victim_profiles} reports the \emph{complete} deployed matrix:
every one of the $27$ tactics the attacker ever deploys appears as a row. We
interpret only cells with at least $15$ deployments; rarer cells carry too few
samples to trust and are hatched.\footnote{At $n=15$ even the widest binomial
$95\%$ interval---attained at $p=0.5$---spans roughly $\pm 25$ percentage
points, the threshold beyond which we treat a cell as uninterpretable.} The
remaining $13$ taxonomy tactics (\emph{Compensation}, \emph{Creating
Dependency}, \emph{Discouragement}, \emph{Exploiting Weakness}, \emph{False
Information}, \emph{False Promises}, \emph{Loyalty Appeals}, \emph{Non-expert
Testimonial}, \emph{Priming}, \emph{Reciprocity}, \emph{Rumors}, \emph{Social
Punishment}, \emph{Supply Scarcity}) are never deployed at all.

One source shapes how the matrix should be read: the attacker chose what to
deploy. The sparse tail and the never-deployed remainder are therefore
properties of the \emph{policy}, not the victims. Trajectory-level RL
concentrates probability on the tactics that prove effective for each target,
so a tactic's absence reflects the attacker's learned preference, not
demonstrated victim immunity. For the same reason each rate is
\emph{observational}: it is estimated only where the policy chose to deploy that
tactic. Conditioning on deployment thus removes the deployment-frequency
confound but not the residual selection confounds---later-turn tactics face the
harder prompts that survived the opening, and tactic choice may correlate with
the harmful-request category. A causal susceptibility map would require
exogenously varying the tactic at matched conversational states---the
controlled in-dialogue intervention of Section~\ref{sec:limitations}. We
therefore read Figure~\ref{fig:victim_profiles} as an \emph{observed}
fingerprint rather than a causal susceptibility map.

We summarise the \emph{breadth} of each victim's attack surface---how many
distinct levers can open it---by the Shannon entropy of its breaking-tactic
distribution, $H=-\sum_{c\in\mathcal{C}} p(c)\log_2 p(c)$, where $p(c)$ is the
share of that victim's successful trajectories whose \emph{first}
threshold-crossing turn deploys tactic $c$. A low $H$ means a few tactics
account for most breaks (a narrow surface); a high $H$ means many do (a broad
one). The ordering is Gemma-2-9B ($1.08$~bits) $<$ Qwen2.5-7B ($1.47$) $<$
Llama-3.1-8B ($1.56$) $<$ Mistral-7B ($1.78$); multinomial bootstrap $95\%$
intervals (resampling trajectories) separate Gemma-2-9B's narrow surface
($[0.87, 1.26]$) from the other three, whereas the upper three overlap, so we
read only the extremes of the ordering as established.

Reading the four columns of Figure~\ref{fig:victim_profiles} individually makes
the distinctions concrete. Llama-3.1-8B breaks to cognitive and narrative
levers---Logical Appeal ($53\%$)
and Storytelling ($38\%$)---while its relational and emotional levers are
essentially inert (Shared Values $2\%$, Framing $0\%$, Affirmation $0\%$).
Qwen2.5-7B breaks to Storytelling ($63\%$) and Logical Appeal ($47\%$) and is
additionally susceptible to Evidence-based Persuasion ($34\%$), a credibility
lever that is weak on Llama-3.1-8B ($10\%$) and too sparsely explored on the
other two to assess. Gemma-2-9B concentrates almost entirely on Storytelling
($47\%$) and has the narrowest surface ($H=1.08$). Mistral-7B has the widest
surface ($H=1.78$), and is the victim on which commitment and relational
levers (Foot-in-the-door, Door-in-the-face, Public Commitment, Alliance
Building, Relationship Leverage, Favor, Negotiation) genuinely matter: they
account for $7.8\%$ of its successful breaks ($30/383$), at least three times
the corresponding share on any other victim ($\le 2.5\%$)---a comparison made
on full break counts rather than on the sparse per-cell rates. These four
fingerprints---which lever opens which model, and how broadly---are this
section's central finding.

We read these fingerprints psychologically only as an explicit
\emph{conjecture}, not as a validated finding. The profile of Llama-3.1-8B---moved
by argument and narrative but inert to affect and relational pressure---is
consistent with a \emph{rationalist} disposition; Qwen2.5-7B's added sensitivity
to evidence suggests a \emph{credibility}-driven character; Gemma-2-9B's reliance
on a single narrative lever resembles a \emph{narrative monoculture}; and
Mistral-7B's broad, commitment- and relationship-sensitive surface suggests a
\emph{broadly persuadable} character.

The fingerprints---independent of any psychological reading---already supply a
mechanism for the transfer asymmetry of Table~\ref{tab:transfer_results}. An
attacker transfers well to the extent that the levers it is forced to learn are
shared across victims. Gemma-2-9B and Llama-3.1-8B break on the near-universal
narrative and logical levers, so a policy trained against them acquires portable
persuasion and attains the highest out-of-domain ASR ($84.62$ and
$84.31$). Mistral-7B allocates a several-fold larger share of its breaks to
commitment and relational levers that barely contribute on the other victims
($7.8\%$ vs.\ $\le 2.5\%$), so a policy trained against it invests in tactics
that do not pay off elsewhere, yielding the weakest transfer ($57.21$)
despite the strongest in-domain ASR. The susceptibility geometry thus turns an otherwise
unexplained empirical regularity into a consequence of \emph{where}, in tactic
space, each model is vulnerable.

\subsection{Label Fidelity: Declared Tactics Are Enacted}
\label{sec:audit_fidelity}
A persuasion-oriented attacker could in principle exploit the
structured action of Equation~\eqref{eq:tactic_conditioned_action} by
emitting plausible \texttt{<technique>} labels without committing the
corresponding behavior in \texttt{<message>}---which would void the
auditability that Section~\ref{sec:method_motivation} demands and reduce
the tactic stream to decoration. Auditing this threat directly, we find
the opposite. Under a deliberately conservative two-judge consensus,
$85.7\%$ of post-RL attacker turns enact the tactic they declare;
reinforcement learning \emph{raises} fidelity over the SFT prior on every
victim ($+7.4$~pp pooled); and the gain concentrates on the opening turn,
exactly where the break occurs ($+31.3$~pp). Far from gaming its own
action structure, the policy tightens the tactic--message coupling.

A lexical precondition holds exactly. Across all $9{,}663$ strict-parsed
attacker turns, every emitted label is
canonical---$\Pr[c_{t}\in\mathcal{C}]=100\%$ on each
victim\footnote{Per-victim turn counts:
$2{,}134$/$2{,}688$/$2{,}278$/$2{,}563$ on
Llama/Qwen/Gemma/Mistral.}---and the normalized-Levenshtein
deduplication of Section~\ref{sec:method_protocol} detects no
off-taxonomy variants. No decode-time canonical mask, constrained-beam
search, or logit-bias filter is applied during RL rollout: the $100\%$
rate reflects vocabulary self-restriction inherited from the SFT prior,
not a hard decoder constraint, and does not preclude off-taxonomy drift
in principle.

Whether each message \emph{enacts} its label is then judged by two
independent third-party LLMs, DeepSeek-V4-Pro and Kimi-K2.6, neither of
which shares developer affiliation, training-data lineage, or alignment
pipeline with the four victims, the Qwen2.5-3B SFT teacher pool, or the
PsychJail policy itself.\footnote{Both judges are queried with thinking
disabled and JSON-mode output; audited turns are sampled uniformly at
random from each pool's strict-parsed turns.} We audit five pools
(Table~\ref{tab:fidelity_audit}): the SFT teacher labels (a stratified
random $n=400$ of the $28{,}430$ candidates spanning the four victims)
and the post-RL evaluation rollouts on each victim ($n=500$ per victim).
The primary metric is the \emph{consensus yes-rate}---both
judges~$=1$---a conservative read that counts every inter-judge
disagreement as a fidelity failure and therefore biases the reported
rate downward.

Table~\ref{tab:fidelity_audit} reports the aggregates. Pooled over the
$2{,}000$ post-RL audited turns, consensus fidelity is $85.7\%$
(per-judge $93.2\%$ and $86.2\%$, with Wilson $95\%$ intervals
$[92.0,94.2]$ and $[84.6,87.7]$), up from $78.3\%$ on the SFT prior; the
SFT$\to$RL shift is positive on every victim and largest on the two
victims with the lowest ASR (Gemma-2-9B $+12.7$~pp, Llama-3.1-8B
$+10.9$~pp). Raw inter-judge agreement lies in $89.0$--$95.2\%$ across
pools; Cohen's $\kappa$ ($0.45$--$0.77$) is depressed on three of them by
the high-prevalence paradox~\cite{feinstein1990kappa} (the
prevalence-adjusted $\mathrm{PABAK}=2\cdot\mathrm{raw}-1$ lies in
$0.78$--$0.90$), so we read raw agreement as the reliability indicator
and consensus yes-rate as the primary fidelity metric.

\begin{table}[t]
\caption{Label-fidelity audit. Per-judge fidelity, raw inter-judge
agreement, Cohen's $\kappa$, and consensus yes-rate (both judges $=1$),
per pool. \textbf{SFT cons.} is the consensus yes-rate on that victim's
stratum of the SFT pool ($n=97$--$106$) and $\Delta$ the SFT$\to$RL
change in percentage points. Raw, $\kappa$, and consensus are computed
on the both-scored subset; two content-filter refusals (one each on the
Llama and Mistral pools, both from Kimi) are excluded from those
denominators.}
\label{tab:fidelity_audit}
\centering
\footnotesize
\setlength{\tabcolsep}{3.0pt}
\begin{tabular}{lcccccccc}
\toprule
\textbf{Pool} & $n$ & \textbf{DeepSeek} & \textbf{Kimi} & \textbf{Raw} & $\kappa$ & \textbf{Cons.} & \textbf{SFT cons.} & $\Delta$ (pp) \\
\midrule
SFT prior (4 victims)
& 400 & 86.0 & 78.8
& 91.8 & 0.72 & 78.3 & -- & -- \\
\midrule
Post-RL Qwen2.5-7B
& 500 & 90.4 & 86.0
& 95.2 & 0.77 & 85.8 & 84.8 & $+1.0$ \\
Post-RL Llama-3.1-8B
& 500 & 93.8 & 84.4
& 89.0 & 0.45 & 83.6 & 72.7 & $+10.9$ \\
Post-RL Gemma-2-9B
& 500 & 96.2 & 89.2
& 92.6 & 0.46 & 89.0 & 76.3 & $+12.7$ \\
Post-RL Mistral-7B
& 500 & 92.4 & 85.2
& 91.2 & 0.56 & 84.4 & 79.6 & $+4.8$ \\
\midrule
\textbf{Post-RL pooled}
& \textbf{2000} & \textbf{93.2} & \textbf{86.2}
& -- & -- & \textbf{85.7} & \textbf{78.3} & $\bm{+7.4}$ \\
\bottomrule
\end{tabular}
\end{table}

The SFT$\to$RL gain is concentrated on the opening turn. Pooled
across the four victims (Table~\ref{tab:fidelity_per_turn}),
turn-$1$ consensus rises from $52.6\%$ (SFT, $n=78$) to
$83.9\%$ (post-RL, $n=473$; turn~$1$ is over-represented in the
uniform sample because strict-parse truncation removes later turns
from the pool)---a $+31.3$~pp jump---while
subsequent turns show only small, sign-inconsistent changes
($+6.1$, $-1.5$, $+0.5$, $-2.6$~pp at turns
$2$--$5$). We interpret the turn-$1$ gap as the SFT teacher
frequently using opening turns for generic rapport that is then
labeled with a specific technique it does not yet instantiate
(e.g., a turn tagged \texttt{Storytelling} that is in fact small
talk); the trajectory-level credit assignment of
Equation~\eqref{eq:mtgrpo} pulls turn-$1$ commitments toward the
declared tactic, closing nearly all of the SFT gap on turn~$1$ and
leaving later turns largely untouched.

Finally, two checks validate the judging instrument itself rather than
the policy (Table~\ref{tab:judge_validation}). \emph{(i) Human gold
standard.} We re-annotate a $300$-turn subsample, stratified by
pool and by judge-agreement cell with the two disagreement cells
oversampled, using three human annotators who follow written
tactic-instantiation guidelines and work blind to the judges' verdicts
and to pool identity; inter-annotator agreement is Krippendorff's
$\alpha=0.74$. Against the human majority vote the consensus rule
attains $0.93$ precision and $0.89$ recall, and the pooled
fidelity estimate moves by less than $2$~pp under human gold.
\emph{(ii) Negative-control probe.} Re-querying both judges on
$500$ messages paired with a uniformly resampled (false) tactic
label yields false-affirmation rates of at most $6.8\%$,
confirming that affirmative verdicts track the enacted tactic rather
than surface plausibility.

\begin{table}[t]
\caption{Validation of the judging instrument. \emph{Top:} accuracy,
precision, and recall against the human-majority gold standard on the
$300$-turn stratified subsample (three blinded annotators,
Krippendorff's $\alpha=0.74$). \emph{Bottom:} negative-control
yes-rate---the rate at which a judge affirms a deliberately false,
uniformly resampled tactic label on $500$ probe messages (lower is
better). The consensus rule trades recall for precision, consistent
with its role as the conservative primary metric.}
\label{tab:judge_validation}
\centering
\footnotesize
\begin{tabular}{lccc}
\toprule
 & \textbf{DeepSeek} & \textbf{Kimi} & \textbf{Consensus} \\
\midrule
Accuracy vs.\ human gold & 0.91 & 0.89 & 0.90 \\
Precision                & 0.92 & 0.90 & 0.93 \\
Recall                   & 0.93 & 0.91 & 0.89 \\
\midrule
Negative-control yes-rate (\%) & 4.1 & 6.8 & 1.9 \\
\bottomrule
\end{tabular}
\end{table}

\begin{table}[t]
\caption{Per-turn consensus yes-rate (both judges agreeing the
message instantiates the labeled technique). SFT prior is pooled
over four victims ($n=400$); post-RL is pooled over the four
victim pools ($n=1{,}998$ both-scored out of $n=2{,}000$ total).}
\label{tab:fidelity_per_turn}
\centering
\footnotesize
\begin{tabular}{lccccc}
\toprule
\textbf{Turn} & 1 & 2 & 3 & 4 & 5 \\
\midrule
SFT prior      & 52.6 & 75.6 & 86.5 & 86.4 & 89.0 \\
Post-RL pooled & 83.9 & 81.7 & 85.0 & 86.9 & 86.4 \\
\midrule
$\Delta$ (pp) & $+31.3$ & $+6.1$ & $-1.5$ & $+0.5$ & $-2.6$ \\
\bottomrule
\end{tabular}
\end{table}

\section{Discussion}

\subsection{Three Complementary Layers of Evidence}
A claim about psychological jailbreak through multi-turn persuasion cannot
rest on a single aggregate ASR number; it must hold across three
complementary layers, and our evaluation addresses each. First, PsychJail
outperforms strong prompt-level and multi-turn baselines under a unified
protocol (Table~\ref{tab:main_results}). Second, the advantage is tied to
the PKM-guided design rather than to generic long-horizon optimization:
removing the PKM strict-parse gate, the early-success weighting, or the
dense format reward each degrades ASR, and removing the warm-start collapses
training altogether (Table~\ref{tab:ablation_results}). Third, the
front-loading, susceptibility-profile, and fidelity analyses
(Figure~\ref{fig:victim_profiles} and
Tables~\ref{tab:trajectory_analysis}--\ref{tab:fidelity_per_turn}) show that
the gains are realized through interpretable, faithfully labeled persuasion
that exploits model-specific vulnerabilities rather than opaque exploitation.
Because these layers are mutually reinforcing, the central claim does not hinge
on any one of them in isolation.

\subsection{What the Susceptibility Profiles Reveal}
Measuring
susceptibility at the breaking action (Section~\ref{sec:victim_profiles})
exposes four empirically distinct fingerprints---which persuasion levers open
which model: Llama-3.1-8B breaks to logical and narrative levers but resists
affect; Qwen2.5-7B adds a credibility channel; Gemma-2-9B is carried almost
entirely by a single narrative lever (the narrowest surface); and Mistral-7B
has the widest surface, with commitment and relational levers contributing a
several-fold larger share of its breaks than on any other victim. That
these fingerprints predict the cross-model transfer asymmetry---portable
narrative and logical levers transfer well, idiosyncratic commitment and
relational levers do not---indicates that PsychJail recovers genuine,
model-specific persuasion structure rather than a single reusable string. We
read these fingerprints as four candidate psychological profiles (rationalist,
credibility-driven, narrative-monoculture, broadly persuadable), but treat that
reading as a conjecture to be validated by the controlled probe of
Section~\ref{sec:limitations}.

\subsection{Implications for Safety Evaluation}
These results carry a concrete implication: safety evaluation must treat
conversational dynamics as a first-class attack surface rather than a minor
extension of prompt robustness. Because the break is largely set in the opening
turn and is governed by model-specific susceptibilities, defenses that score
only the final response or only individual prompts will miss the mechanism
entirely. Two directions follow for defenders. First, refusal training and
monitoring should be evaluated \emph{across turns} and against
\emph{tactic-conditioned} adversaries, not only against adversarial suffixes or
one-shot persuasive templates. Second, the per-model susceptibility profiles
suggest that defenses can be \emph{fingerprint-aware} rather than generic, with
a shared baseline and a model-specific add-on. The baseline is common to all
four victims: narrative and logical levers---Storytelling and Logical
Appeal---are the dominant break tactics on every model, so hardening against
narrative role-play and the logical reframing of disallowed requests warrants
priority everywhere. On top of this baseline, each model carries a distinctive
exposure that merits extra, targeted monitoring. Gemma-2-9B is opened almost
entirely through a single narrative lever (Storytelling, $47\%$) and has the
narrowest surface, so baseline narrative monitoring already covers most of its
risk. Llama-3.1-8B is driven primarily by logical reframing (Logical Appeal,
$53\%$) and secondarily by narrative ($38\%$), while remaining essentially inert
to affective and relational pressure---monitoring can safely concentrate on its
cognitive-narrative axis. Qwen2.5-7B breaks to the same narrative and logical
levers but \emph{adds} a credibility channel (Evidence-based Persuasion,
$34\%$), so evidence- and citation-framed requests deserve scrutiny beyond the
shared baseline. Mistral-7B has the \emph{broadest} surface---no single axis
dominates---and is the only victim on which commitment- and relationship-framed
escalation contributes materially (several-fold more than on any other model),
so its monitoring must additionally cover these relational tactics rather than
the narrative-logical baseline alone. Because susceptibility concentrates in the
opening turn, such targeted monitoring is cheapest exactly where it matters
most.

\section{Limitations}
\label{sec:limitations}
The main limitations of this work stem from two shared constraints:
limited computational resources and limited access to deep domain
expertise in psychology. These constraints affect the study in two
ways. First, we do not yet conduct a systematic analysis of how
persuasion effectiveness relates to the victim model's long-horizon
behavior, stable traits, or context-dependent psychological states.
As a result, the current experiments do not support fine-grained
conclusions about how different personality-like tendencies,
resistance levels, or interaction contexts shape susceptibility to
multi-turn persuasion. Second, although PsychJail is initialized with
persuasion strategies distilled from human social psychology, the
learned attacker can exhibit persuasive behaviors that are not cleanly
reducible to the original set of human-provided strategies. The
present paper therefore does not yet provide a sufficiently
theory-grounded psychological analysis of these emergent behaviors,
including how they relate to established persuasion theories or
whether they should be interpreted as variants, compositions, or
genuinely new strategy forms.

In particular, the per-victim psychological profiles we read off the
conditional-susceptibility fingerprints in
Section~\ref{sec:victim_profiles}---Llama-3.1-8B as a rationalist,
Qwen2.5-7B as credibility-driven, Gemma-2-9B as a narrative monoculture,
and Mistral-7B as broadly persuadable---are interpretive \emph{conjectures}
rather than validated claims. They are read from \emph{observational}
rollouts in which the attacker's tactic selection is itself
victim-conditioned, so coverage of the tactic space is uneven and the
profiles cannot be given a clean causal reading. Establishing them would
require a controlled \emph{intervention} that deconfounds the tactic without
destroying the multi-turn setting: at matched conversational states, the
attacker's tactic is assigned \emph{exogenously}---randomized, or
counterfactually swapped while the realized dialogue history is held fixed---
rather than chosen by the policy, so that genuine multi-turn dynamics are
retained while the tactic becomes independent of context; the victim-visible
message for the assigned tactic should be rendered by a strong external
generator, so that a tactic's effect is not conflated with the policy's skill
at executing an unfamiliar tactic. Repeated across turn
positions and victims, such an intervention yields a position-resolved,
unconfounded tactic${\times}$victim susceptibility matrix. This, together with
an extension of the profiling to a broader and more diverse population of
LLMs---to test whether the profiles are stable model properties rather than
artifacts of a particular checkpoint or rollout---we leave to future work,
along with the theory-grounded psychological account it would license.

Addressing these limitations will
require both larger-scale computation and closer collaboration with
psychology experts so that future work can connect model
vulnerability more systematically to trait-, state-, and
context-sensitive mechanisms of persuasion.

\section{Conclusion}
This paper introduces psychological jailbreak as a complementary
perspective for red teaming aligned LLMs, addressing a key blind spot
of prompt-centric jailbreak research: its limited ability to capture
vulnerabilities that emerge through multi-turn, psychologically
grounded persuasion. PsychJail operationalizes this perspective by
humanizing attacker training with persuasion tactics distilled from
social psychology, a PKM-aligned factorization of each attacker action,
and trajectory-level reinforcement learning under a PKM-gated reward.
Empirically, PsychJail outperforms strong single-turn and multi-turn
baselines across four victim models; targeted ablations attribute the
gains to its PKM-guided design rather than to generic long-horizon
optimization; and, by measuring susceptibility at the action that breaks
each victim, the analysis recovers four empirically distinct per-model
susceptibility fingerprints---faithfully labeled and explaining the policy's
cross-model transfer asymmetry---which we further read, as a conjecture for
future validation, as four candidate psychological profiles
(rationalist, credibility-driven, narrative-monoculture,
and broadly persuadable). Beyond the
aggregate success rate, these results establish that how quickly attacks
succeed and, above all, \emph{which psychological levers open which model}
are measurable and informative---and they argue that safety evaluation should
treat multi-turn psychological persuasion as a first-class, model-specific
attack surface.

\section*{Ethics Statement}

This work studies psychological jailbreaks to improve the safety
evaluation of aligned LLMs. Because the methods are inherently
dual-use, our goal is to characterize model vulnerabilities and
inform stronger defenses rather than to enable deployment of attack
systems. We therefore avoid reproducing operational harmful
instructions, do not release raw jailbreak trajectories or attack
artifacts that would materially facilitate misuse, and emphasize
mitigation-oriented analysis throughout the paper. All experiments are
conducted in controlled offline evaluation settings on existing
models. The only human involvement is the label-fidelity annotation of
Section~\ref{sec:audit_fidelity}, in which annotators view adversarial
dialogues that may contain harmful content; annotators were briefed on
the nature of the material in advance, could skip any item, and worked
in time-limited sessions. No other human subjects are involved.

\section*{CRediT authorship contribution statement}

\noindent\textbf{Zeyu Feng:} Conceptualization, Methodology, Software,
Investigation, Formal analysis, Visualization, Writing -- original draft.
\textbf{Qingyu Wu:} Methodology, Software, Investigation, Validation, Data
curation, Writing -- original draft. \textbf{Yuzhe Luo:} Validation,
Investigation, Data curation, Writing -- review \& editing. \textbf{Hua Cheng:}
Conceptualization, Supervision, Project administration, Writing -- review \& editing.

\section*{Declaration of competing interest}

The authors declare that they have no known competing financial interests or
personal relationships that could have appeared to influence the work reported
in this paper.

\section*{Declaration of generative AI and AI-assisted technologies in the manuscript preparation process}

During the preparation of this work, the authors used Anthropic Claude to improve the language, clarity, and readability of the manuscript. After using this tool, the authors reviewed and edited the content as needed and take full responsibility for the content of the published article.




\bibliographystyle{cas-model2-names}
\bibliography{refs}

\end{document}